\documentclass[pdflatex,sn-mathphys-num]{sn-jnl}

\usepackage{threeparttable}
\usepackage[title]{appendix}%
\usepackage{xcolor}%
\usepackage{textcomp}%
\usepackage{manyfoot}%
\usepackage{booktabs}%
\usepackage{comment}
\usepackage{lmodern}
\usepackage{graphicx}%
\usepackage{multirow}%
\usepackage{amsmath,amssymb,amsfonts}%
\usepackage{amsthm}%
\usepackage{mathrsfs}%
\usepackage{float}
\usepackage{xcolor}

\theoremstyle{thmstyleone}%
\theoremstyle{thmstyletwo}%

\theoremstyle{thmstylethree}%

\begin{document}


\title[Article Title]{Identification of fixations and saccades in eye-tracking data using adaptive threshold-based methods}



\author[1]{\fnm{Charles} \sur{Orioma}}
\equalcont{First two authors contributed equally, developing the analysis. Last two authors contributed equally, coordinating and supervising the work.}

\author[1]{\fnm{Josef} \sur{Krivan}}
\equalcont{First two authors contributed equally, developing the analysis. Last two authors contributed equally, coordinating and supervising the work.}

\author[1]{\fnm{Rujeena} \sur{Mathema}}

\author[1,2]{\fnm{Pedro} \sur{Lencastre}}

\author[1,2,3]{\fnm{Pedro G.} \sur{Lind}}

\author[4]{\fnm{Alexander} \sur{Szorkovszky}}
\equalcont{First two authors contributed equally, developing the analysis. Last two authors contributed equally, coordinating and supervising the work.}

\author*[1,2]{\fnm{Shailendra} \sur{Bhandari}}\email{shailendra.bhandari@oslomet.no}
\equalcont{First two authors contributed equally, developing the analysis. Last two authors contributed equally, coordinating and supervising the work.}

\affil*[1]{\orgdiv{Department of Computer Science}, \orgname{OsloMet -- Oslo Metropolitan University}, \orgaddress{\street{P.O.~Box 4 St.~Olavs plass}, \city{Oslo}, \postcode{N-0130}, \country{Norway}}}

\affil[2]{\orgdiv{OsloMet Artificial Intelligence Lab}, \orgname{OsloMet -- Oslo Metropolitan University}, \orgaddress{\street{Pilestredet 52}, \city{Oslo}, \postcode{N-0166}, \country{Norway}}}

\affil[3]{\orgdiv{School of Economics, Innovation and Technology}, \orgname{Kristiania University of Applied Sciences}, \orgaddress{\street{Kirkegata 24-26}, \city{Oslo}, \postcode{N-0153}, \country{Norway}}}

\affil[4]{\orgdiv{Numerical Analysis and Scientific Computing}, \orgname{Simula Research Laboratory}, \orgaddress{\city{Oslo}, \postcode{N-0164}, \country{Norway}}}

\abstract{%
Properties of ocular fixations and saccades are highly stochastic during many experimental tasks, and their statistics are often used as proxies for various aspects of cognition. Although distinguishing saccades from fixations is not trivial, experimentalists generally use common ad-hoc thresholds in detection algorithms. This neglects inter-task and inter-individual variability in oculomotor dynamics, and potentially biases the resulting statistics. In this article, we introduce and evaluate an adaptive method based on a Markovian approximation of eye-gaze dynamics, using saccades and fixations as states such that the optimal threshold minimizes state transitions. Applying this to three common threshold-based algorithms (velocity, angular velocity and dispersion), we evaluate the overall accuracy against a multi-threshold benchmark as well as robustness to noise. We find that a velocity threshold achieves the highest baseline accuracy (90-93\%) across both free-viewing and visual search tasks. However, velocity-based methods degrade rapidly under noise when thresholds remain fixed, with accuracy falling below 20\% at high noise levels. Adaptive threshold optimization via $K$-ratio minimization substantially improves performance under noisy conditions for all algorithms. Adaptive dispersion thresholds demonstrate superior noise robustness, maintaining accuracy above 81\% even at extreme noise levels ($\sigma = 50$\,px), though a precision-recall trade-off emerges that favors fixation detection at the expense of saccade identification. In addition to demonstrating our parsimonious adaptive thresholding method, these findings provide practical guidance for selecting and tuning classification algorithms based on data quality and analytical priorities.
}




\maketitle

\section{Introduction}\label{sec1}

The human visual system orchestrates a sophisticated interplay of eye movements, fixations, saccades, smooth pursuits, and vergence—that reveals intricate patterns of attention, cognition, and neurological function \citep{Duchowski2017}. Eye-tracking technology, utilizing corneal reflections and pupil dynamics at sampling rates often exceeding 1,000 Hz, captures these movements with high precision, mapping foveal gaze to elucidate visual perception \citep{68d3bfb814cc4a61aae0131dd135c9ec}. Fixations, stable gazes lasting 200–400 ms to process detailed visual information, and saccades, rapid eye movements reaching velocities up to 500$^\circ$/s, form the cornerstone of applications across diverse fields, including psychology, neuroscience, clinical diagnostics, education, marketing, and human-computer interaction \citep{10.1145/355017.355028}. From early diagnosis of neurological disorders such as Alzheimer’s and schizophrenia \citep{korda2018automatic}, to enhancing educational tools through gaze-based learning analytics \citep{busjahn2014eye}, to optimizing consumer behavior analysis in visual marketing \citep{MKT-011}, eye-tracking has transformative potential. Yet, accurately distinguishing fixations from saccades remains a central challenge, as traditional algorithms struggle with noise, blinks, and behavioral variability across diverse visual stimuli \citep{schweitzer2020adaptive}.

Conventional algorithms, such as Velocity-Threshold Identification (I-VT) and Dispersion-Threshold Identification (I-DT), rely on fixed velocity (e.g., 20–40°/s for I-VT) or spatial dispersion thresholds to classify eye movements \citep{10.1145/355017.355028}. While computationally efficient and suitable for real-time applications, these methods are sensitive to noise and often fail to detect auxiliary events like microsaccades or post-saccadic oscillations (glissades), limiting their robustness in complex datasets \citep{martinezs, nystrom2010adaptive}. Recent advancements have introduced more sophisticated approaches to address these limitations. For instance, Larsson et al. developed robust segmentation techniques for fixations and smooth pursuits in high-speed eye-tracking data, outperforming velocity- and dispersion-based methods \citep{LARSSON2015145}. Zemblys et al. leveraged machine learning, specifically random forest classifiers, to automate event detection with accuracy comparable to manual coding \citep{zemblys}. Additionally, methods like hidden Markov models (HMM) and minimum spanning trees offer nuanced classification by modeling temporal and spatial gaze patterns \citep{10.1145/355017.355028}. Despite these innovations, traditional algorithms remain prevalent due to their simplicity and versatility, necessitating rigorous evaluation to optimize their performance across varied tasks \citep{10.1145/355017.355028}. Recent comprehensive evaluations have demonstrated that algorithm choice significantly impacts classification outcomes \citep{Andersson2017OneAlgorithm}, motivating the need for systematic comparison studies.

This study evaluates the efficacy of threshold-based algorithms for classifying fixations and saccades in eye-tracking data collected from free viewing and visual search tasks, and introduces an adaptive method for threshold selection based on minimization of state transitions, using a quantity we term the {\em $K$-ratio}. We then perform a Gaussian noise analysis to evaluate robustness (e.g., detecting performance degradation at noise levels \(\geq\) 2 pixels). The analysis focuses on implementation difficulty, computational complexity, usability, and classification effectiveness, addressing the question: ``Which algorithms are best suited for classifying eye-tracking features and movements?'' Our findings aim to provide actionable insights for improving classification accuracy, advancing applications in cognitive research, clinical diagnostics, and assistive technologies.

\section{Data and methods}\label{sec11}

\subsection{``Pure" and ``noisy" eye-gaze trajectories}

Eye movements were recorded at Oslo Metropolitan University in 2023–2024 under a Virtual-Eye project data-sharing agreement \cite{rujeena2025-eye-gaze-dynamics-datasets}. More than 250 adults participated in the broader study; for the present report, we analyzed a matched within-subject subset of 15 participants who completed both tasks with high-quality data. The broader dataset, which includes the present tasks used in this study together with several additional experimental paradigms, has been published online as a data descriptor in \textit{Scientific Data} \citep{rujeena2025-eye-gaze-dynamics-datasets}.

Recordings used the EyeLink Portable Duo (SR Research, Canada) at 1000~Hz in head-stabilized mode with a 9-point calibration \citep{srresearch2024eyelink}, and validation yielded accuracy $\leq 0.5^{\circ}$. Stimuli were presented on an ASUS ROG Swift PG259QNR display (1920$\times$1080, 360~Hz, 1~ms) at a viewing distance of 100~cm with a chin rest. PsychoPy (v2022.2.4) controlled presentation and synchronization with Python (v3.9) logging.

Two paradigms were used for the same 15 participants. In a \emph{free-viewing} condition (Random Pixel), participants viewed unstructured pixel fields (3 trials, 60~s each; $\sim$180{,}000 samples per participant, $\sim$2.7~M total in this subset). In a \emph{visual search} condition (Waldo), participants located targets in complex scenes from \emph{Where's Waldo?} (9 trials, 45~s each; $\sim$405{,}000 samples per participant, $\sim$6~M total in this subset) \citep{bookwaldo}. Each dataset comprised ten columns per sample at 1000~Hz: timestamp; left and right $x,y$ gaze in pixels; left and right pupil size; and event flags. Blinks and tracking loss are handled by removing samples with missing gaze coordinates (NaNs) before all feature computations. See Table~\ref{tab:dataset-comparison}.
\begin{table}[t]
    \centering
    \caption{Comparison of random pixel and Waldo task datasets.}
    \label{tab:dataset-comparison}
    \begin{tabular}{lcc}
        \hline
        \textbf{Characteristic} & \textbf{Random pixel task} & \textbf{Waldo task} \\
        \hline
        Number of participants & 15 & 15 \\
        Data points per participant & $\sim$180,000 & $\sim$405,000 \\
        Total data points & 2,700,252 & $\sim$6,000,000 \\
        Row count range & 180,003--180,044 & 405,088--405,163 \\
        Number of columns & 10 & 10 \\
        Sampling rate & 1000 Hz & 1000 Hz \\
        Approximate duration & 3 minutes & 6.75 minutes \\
        \hline
    \end{tabular}
\end{table}

Binocular coordination was assessed via Pearson correlation and linear regression on left/right eye X/Y coordinates, with fixation/saccade agreement calculated as percentage overlap. In the Random Pixel dataset, binocular coordination was high: Pearson correlations between left and right eyes were (r=0.9948) for the (x)-coordinate and (r=0.9748) for the (y)-coordinate. Framewise agreement between the two eyes’ event labels was 98.30\% for fixations and 98.34\% for saccades. Across the full datasets, the overall event distribution was dominated by fixations (\(\sim 92\%)\). 
In the rest of this article, we use the left eye only for the X/Y coordinates.

The EyeLink software uses a combination of velocity and acceleration thresholds to classify saccades and fixations. We used the default EyeLink parser settings (30$^\circ$/s velocity, 8000$^\circ$/sec$^2$ acceleration) to extract reference event labels per frame, which serve as a practical benchmark rather than a physiological ground truth.
To study the effect of noise in the classification algorithm, we augmented our data sample by generating a superposition of the collected trajectories with a tunable level of noise.
These ``noisy" trajectories were then analyzed additionally to the collected eye-gaze trajectories.
We restricted to the case of  
Gaussian noise, with $\varepsilon_{x,i}$ and $\varepsilon_{y,i}$, added to each observation $i$, at the location $(x_i,y_i)$: 
\begin{equation}
P^{\text{(noisy)}} = \{(x_i^{\text{(noisy)}},y_i^{\text{(noisy)}} ) \}   = \{(x_i + \varepsilon_{x,i}, y_i + \varepsilon_{y,i})\} \, ,
\end{equation}
where 
$\varepsilon_{x,i}, \varepsilon_{y,i} \sim \mathcal{N}(0, \sigma^2)$ 
are normally distributed with zero mean and a tunable standard deviation $\sigma$. \(\sigma\) should be interpreted primarily as a controlled stress-test parameter for algorithmic robustness rather than a calibrated estimate of typical experimental measurement noise, although the smallest values \((\sigma \approx 1\)–2 px) are in the range of plausible high-quality recordings.

\subsection{Classification algorithms}

We introduce adaptive versions of the three classification algorithms mentioned above, I-VT and I-DT \cite{10.1145/355017.355028}, as well as an angle-weighted variant of I-VT which we call Angular Velocity-Threshold Identification (I-AVT). 
The I-VT algorithm uses the instantaneous velocity $V_i$, which is the velocity calculated between successive measurements, namely:
\begin{equation}
V_i = \frac{\sqrt{(x_{i+1} - x_{i})^2 + (y_{i+1} - y_{i})^2}}{t_{i+1} - t_{i}},
\end{equation}
where \(x_i, y_i\) are coordinates at timestamp \(t_i\). All kinematic features used by the threshold algorithms are computed from the recorded screen coordinates and are therefore expressed in pixel-based units (px/s for velocity-like quantities), unless explicitly converted to degrees of visual angle.

The I-AVT method uses the effective angular velocity, which combines the scalar velocity and the directional persistence, namely
\begin{equation}
V^{\text{eff}}_i = V_i \cdot \cos (\theta_{i}-\theta_{i-1}),
\end{equation}
where $\theta_i$ is the angle of the displacement vector with the positive horizontal axis:
\begin{equation}
\theta_i = \tan^{-1}\left(\frac{y_{i+1}-y_i}{x_{i+1}-x_i}\right) \; .
\end{equation} 
Savitzky-Golay smoothing filter was applied before computing I-AVT to reduce high-frequency fluctuations at 1000~Hz. The cosine term is computed only for nonzero displacement vectors; samples with zero-length steps are excluded, and the effective velocity is taken.

The I-DT algorithm identifies fixations based on the spatial dispersion of gaze points within a moving time window. It uses two parameters, a minimum time window span $T_\mathrm{min}$ and a dispersion threshold $D_{max}$. We set \(T_\mathrm{min}=50\,\mathrm{ms}\), which corresponds to 50 samples at 1000 Hz and serves to suppress spurious ‘fixations’ caused by brief noise-driven clustering or microsaccade-scale interruptions. This value is also well below typical fixation durations (hundreds of ms), so it acts as a conservative lower bound rather than a tuning knob that shapes the main results.
The algorithm starts at the beginning of the time series and searches for the first index $i$ where
\begin{equation}
\exists k>i:\;(t_k-t_i>T_\mathrm{min}) \land (D_i^k < D_\mathrm{max}) \; ,
\label{conditionfordispersion}
\end{equation}
where $D_i^k$ is the 
dispersion  within a time span $[t_i,t_k]$, combining the dimensions of the smallest rectangle that encompasses all recorded gaze locations between $t_i$ and $t_k$:
\begin{equation}
D_i^k = \max_{j\in[i,k]}(x_j) -\min_{j\in[i,k]}(x_j) + \max_{j\in[i,k]}(y_j) - \min_{j\in[i,k]}(y_j) \; .
\end{equation}
All data points from index $i$ up to the largest index $k$ satisfying the condition in equation (\ref{conditionfordispersion}) are then labelled as a fixation. The next iteration of the algorithm then begins the search from index $k+1$.

\subsection{Optimization through a statistical metric} 

In this section we introduce a metric we call $K$-ratio, which compares the empirical number of fixation-saccade transitions observed in an eye-gaze trajectory with the number of transitions between saccades and fixations that would be expected if data points simply alternated randomly between being fixations and saccades:
\begin{equation}\label{k_ratio}
\text{$K$-ratio} = \frac{p_{\mathrm{empirical}_{F \to S}}}{p_\mathrm{independent F \rightarrow S}}.
\end{equation}
\begin{figure}[t]
\centering
\includegraphics[width=\linewidth]{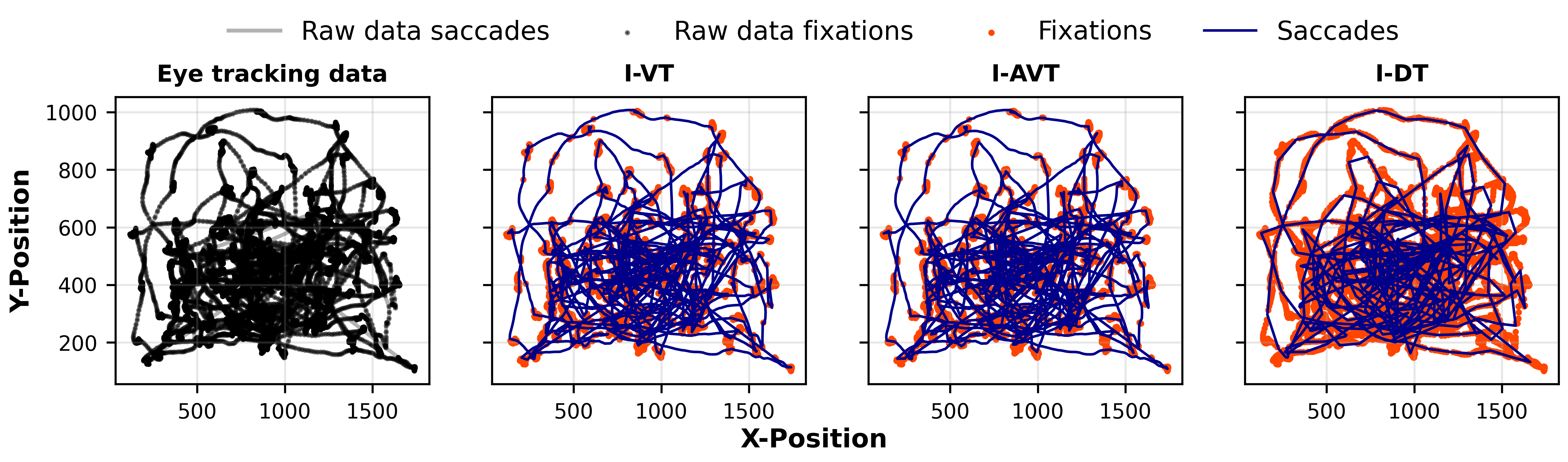}\\
\vspace{8pt}
\includegraphics[width=\linewidth]{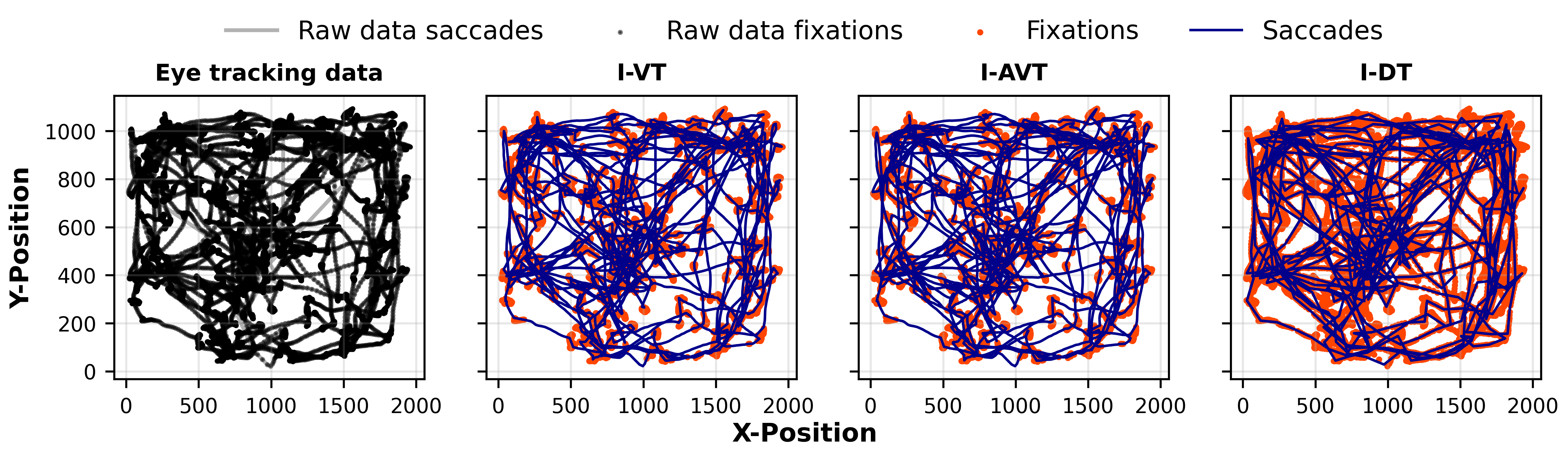}
\caption{Example of scanpaths with overlaid classifications (orange = fixation, blue = saccade). Top row: Random Pixel task; bottom row: Waldo task. Columns (left to right): raw gaze points, I-VT, I-AVT, and I-DT results. All gaze positions are in screen pixels.}
\label{fig:algorithm_comparison}
\end{figure}
If the $K$-ratio is lower than one, it means that the labels tend to agglomerate along the sample, meaning, in our case, that data points labelled as fixations (resp. saccades) tend to follow other data points labelled as fixations (resp. saccades).

If the probability to transition from one state to another is independent of the current state and independent of time, then the expected transition probability from a label $\ell_{\bar{V}_i}(V_j) = F$ to $\ell_{\bar{V}_i}(V_{j+1}) = S$ would be:
\begin{equation}
     p_{\mathrm{independent}_{F \to S}} = Pr \left(\ell_{\bar{V}_i}(V_{j+1}) = S  \right) Pr\left(\ell_{\bar{V}_i}(V_{j})= F \right) = n_S (1-n_S),
 \end{equation}
where $n_S=N_S/N$, with $N_S$ the number of occurrences of the label $S$ and $N$ the total number of observations. However, if the probability of transition from one state to another depends on the current state, this separation is not possible. The $K$-ratio is then
\begin{equation}
   \text{$K$-ratio} =  \frac{ Pr\left(\ell_{\bar{V}_i}(V_{j+1}) = S,\ell_{\bar{V}_i}(V_{j})= F \right) }{Pr\left(\ell_{\bar{V}_i}(V_{j+1}) = S  \right) Pr\left(\ell_{\bar{V}_i}(V_{j})= F \right)}.
\end{equation}
Again, assuming stationarity, this can be written in terms of empirical quantities
\begin{equation}
\text{$K$-ratio} = \frac{n_{F\rightarrow S}}{n_S (1-n_S)}\; ,
\end{equation}
where $n_{F\rightarrow S}$ is the number of fixation-to-saccade transitions divided by the number of data points.
%
We therefore assessed time-local stability of transition statistics and the short-lag adequacy of a first-order Markov proxy by repeating the transition computations in equal-duration blocks and by comparing empirical $k$-step transitions to the Markov prediction $T^k$ (see Appendix A, Fig.~A1). By minimizing this ratio, thresholds used to label fixations and saccades are selected towards keeping these two states separated in time. This approach involves a methodological trade-off: it enhances coherence and suppresses noise-induced artifacts at the expense of sensitivity to high-frequency micro-movements or post-saccadic oscillations. In noisy recordings, this can be beneficial as a form of regularization, but it also implies a trade-off in which short or noise-contaminated saccades may be absorbed into neighboring fixations.

While another kind of state-space modelling (i.e.\ hidden Markov model) has been used to adaptively identify fixations \cite{10.1145/355017.355028}, and velocity-adaptive approaches have been proposed \cite{nystrom2010adaptive}, these require detailed modeling of the velocity distributions of fixations and saccades. In comparison, the $K$-ratio is parsimonious and can be applied to any threshold-based algorithm.

\begin{figure}[t]
\centering
\includegraphics[width=0.95\linewidth]{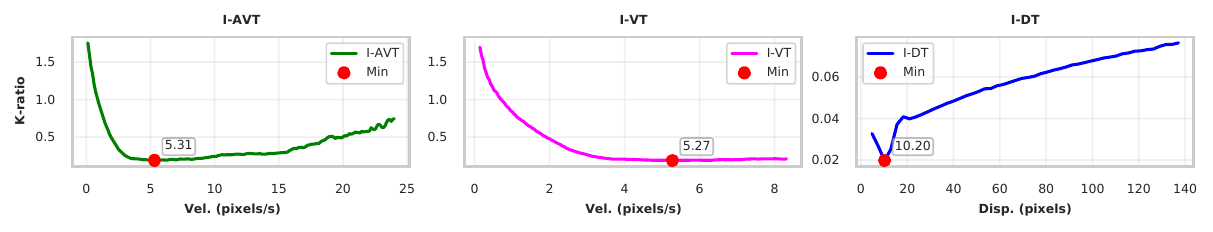}\\
\vspace{4pt}
\includegraphics[width=0.95\linewidth]{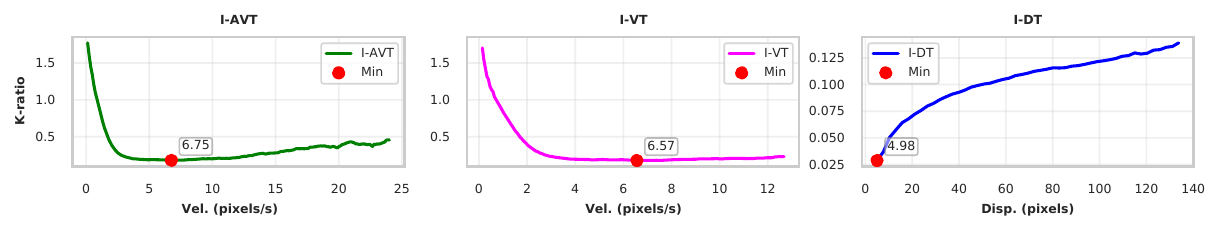}
\caption{$K$-ratio curves used for threshold optimization (top: Random Pixel; bottom: Waldo). Minima define the task-specific thresholds for I-VT, I-AVT, and I-DT. The respective $K$-ratio for each of the algorithms for both task are tabulated in a supplementary section at the end of the paper. See Appendix Table~\ref{tab:k_ratio_all}. For I-DT we use \(T_\mathrm{min}=50\,\mathrm{ms}\). Threshold units are in pixels for I-AVT, I-VT (px/s), and I-DT (px).}
\label{fig:k_ratio_all}
\end{figure}

\section{Results}\label{sec2}

\begin{figure}[t]
\centering
\includegraphics[width=0.49\linewidth]{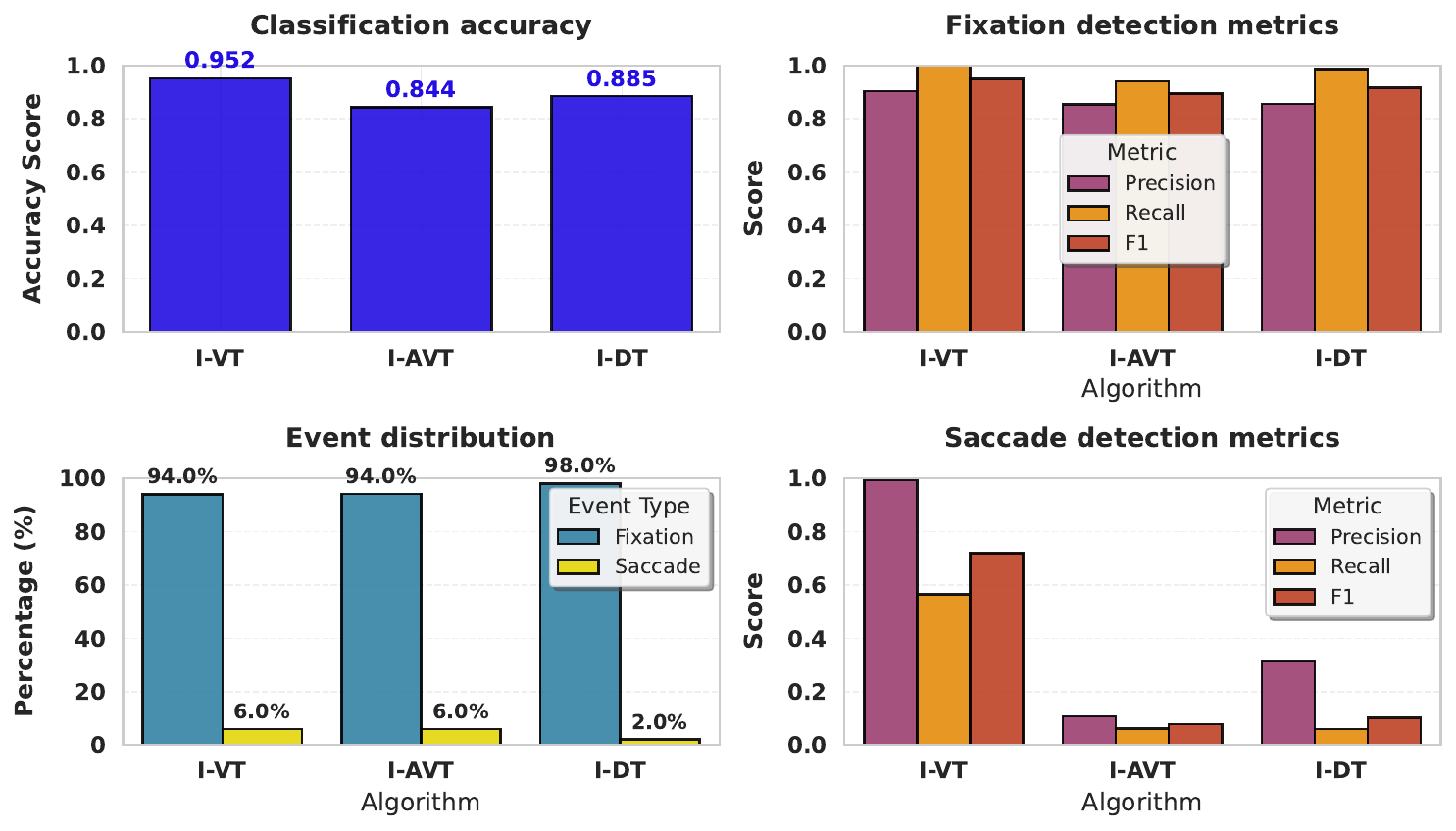}
\hfill
\includegraphics[width=0.49\linewidth]{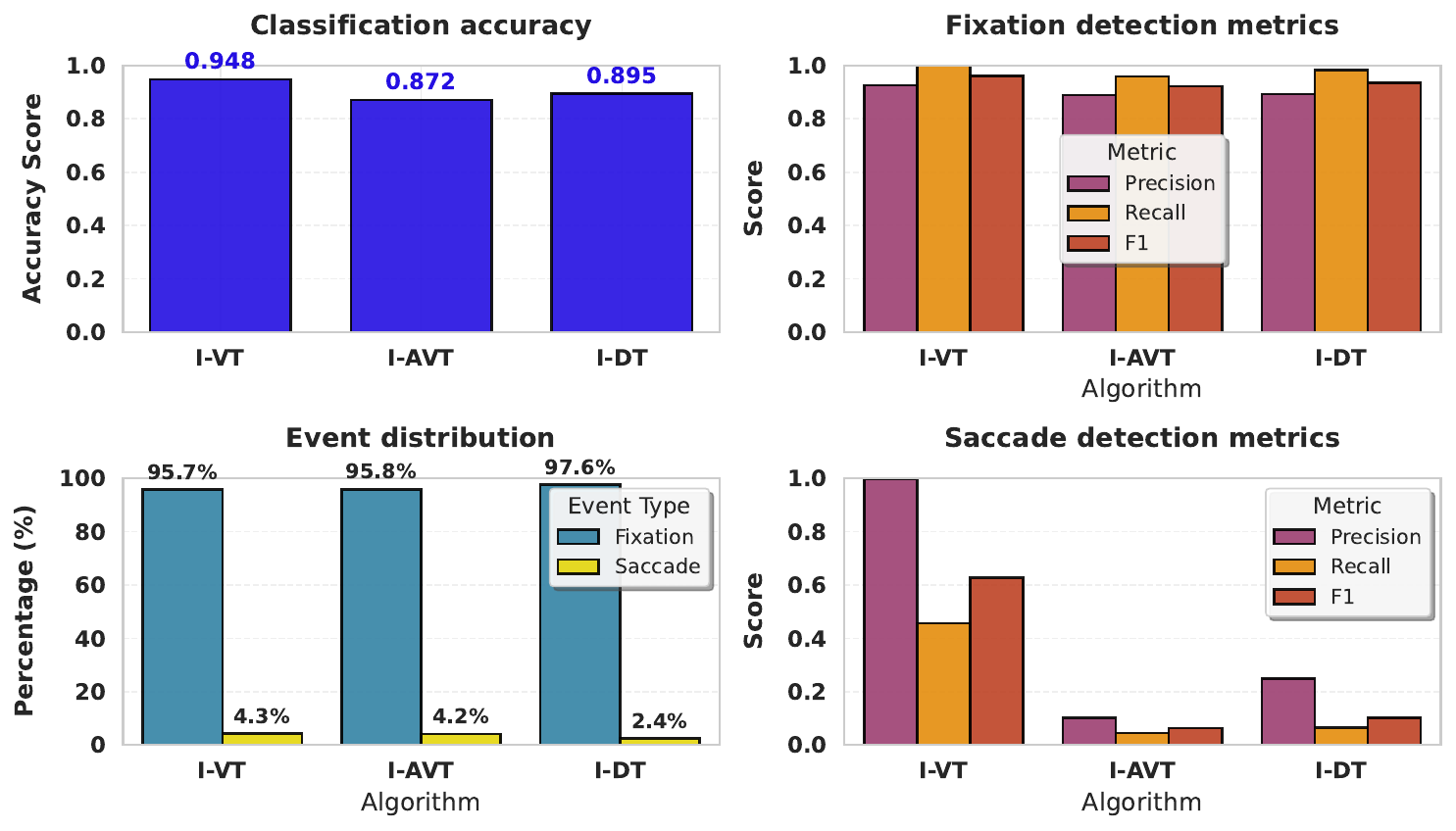}
\caption{Performance metrics at zero noise (left: Random Pixel; right: Waldo) using EyeLink reference labels. Bars display overall accuracy, class-wise precision/recall/F1, and predicted fixation proportion for each algorithm. Since fixation dominates ($\sim92\%$), the saccade F1-score ($\mathrm{F1}$) and its precision/recall components are the most informative summary of rare-event detection performance.}
\label{fig:alg_performance}
\end{figure}

We evaluated I-VT, I-AVT, and I-DT on the Random Pixel (free viewing) and Waldo (visual search) datasets recorded at 1000~Hz, using the EyeLink event labels as frame-wise reference labels. The three classifiers were applied to the same recordings and compared in terms of threshold optimization, frame-wise accuracy, class-wise metrics, and robustness to added Gaussian noise. Frame-wise reference labels were provided by the standard EyeLink parser (30$^{\circ}$/s velocity; 8000$^{\circ}$/s$^{2}$ acceleration).

Figure~\ref{fig:algorithm_comparison} shows representative scanpaths with overlaid classifications for both tasks. I-VT produces compact fixation clusters with clearly delineated saccade segments. I-AVT tends to generate long fixation runs with relatively few saccade segments, while I-DT yields the most fixation-dominated segmentation, often merging brief saccades into extended fixation episodes. These qualitative patterns reflect the underlying class distributions and confusion trends that appear in the aggregate metrics. Thresholds were selected by minimizing the $K$-ratio for each algorithm and task. The $K$-ratio curves for the Random Pixel and Waldo datasets are shown in Fig.~\ref{fig:k_ratio_all}, and the corresponding optimal thresholds and $K$-ratio values are summarized in Appendix Table~\ref{tab:k_ratio_all}. In both tasks, the $K$-ratio decreases sharply at low thresholds and then levels off into a broad minimum. The Waldo task exhibits wider minima than the Random Pixel task, consistent with more heterogeneous oculomotor dynamics during structured visual search.

Overall frame-wise accuracy and class-wise precision, recall, and F1-scores for fixations and saccades are summarized visually in Fig.~\ref{fig:alg_performance}. Because the datasets are strongly fixation-dominated ($\sim$92\%), overall accuracy is driven primarily by fixation performance and is therefore secondary for model comparison. We treat the saccade F1-score ($\mathrm{F1}_S$) as the primary metric, and interpret accuracy only as a contextual summary. Under noise-free conditions, I-VT achieves the highest $\mathrm{F1}_S$ among the three methods (Random Pixel: $\mathrm{F1}_S=0.51$; Waldo: $\mathrm{F1}_S=0.54$), compared with I-AVT (0.46, 0.53) and I-DT (0.42, 0.48) (Table~\ref{tab:merged_ivt}~-~\ref{tab:merged_idt}). Under increasing noise, all methods show declining $\mathrm{F1}_S$, with adaptive thresholding partially mitigating the degradation, but with a pronounced trade-off for I-DT where fixation performance remains high while saccade detection becomes conservative at high $\sigma$ (Table~\ref{tab:merged_idt} and Fig.~\ref{fig:f1_comparison}). All algorithms slightly over-classify fixations
compared to the reference label proportion of 92\%. Across both tasks, I-VT generally showed strong rare-event performance while maintaining high fixation performance. I-AVT maintains high fixation performance but under-detects saccades, leading to low saccade F1-scores despite its use of directional information. I-DT shows a pronounced fixation bias, with the highest fixation proportions and low saccade recall, reflecting the tendency of the dispersion criterion to group short saccades into longer fixation segments. Taken together, the results indicate that I-VT provides a favorable overall balance among the three methods when evaluated against the EyeLink labels, although the relative differences between methods depend on task and noise condition. Both the optimal thresholds and obtained $K$-ratio minima showed significant inter-participant differences. The adaptive I-VT algorithm in particular showed consistently high intraclass correlations, with ICC(1,1) values of 0.688 and 0.728 for the Random Pixel and Waldo task thresholds, respectively, and 0.816 and 0.853, respectively, for their $K$-ratios (see Appendix \ref{app:icc}).

\begin{figure}[t]
\centering
\includegraphics[width=\linewidth]{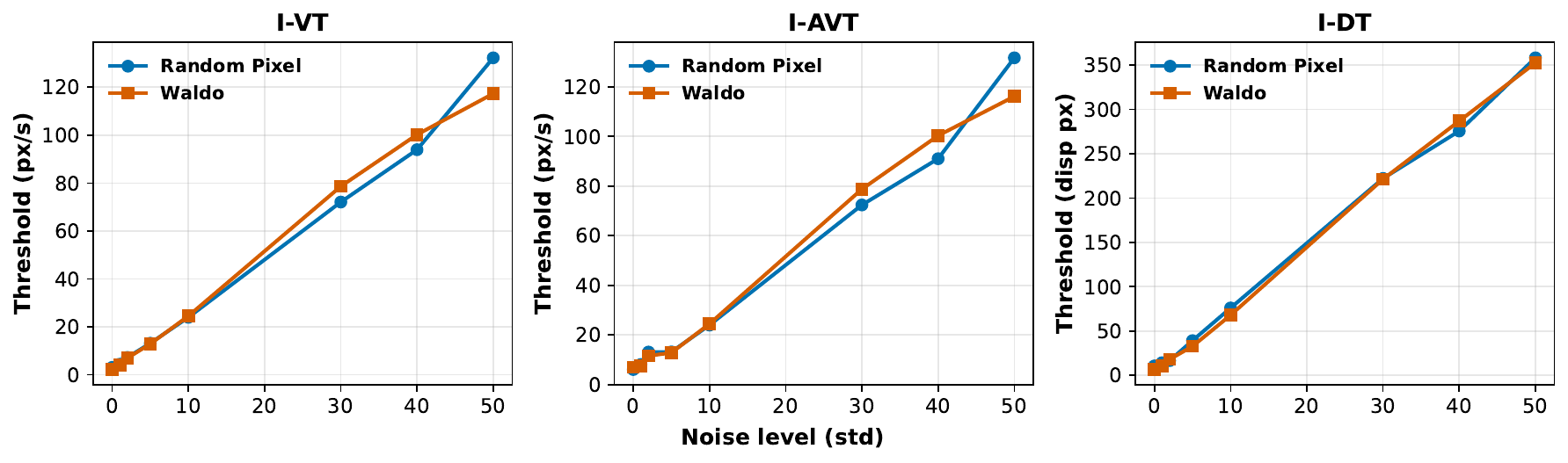}
\vspace{16pt}
\includegraphics[width=\linewidth]
{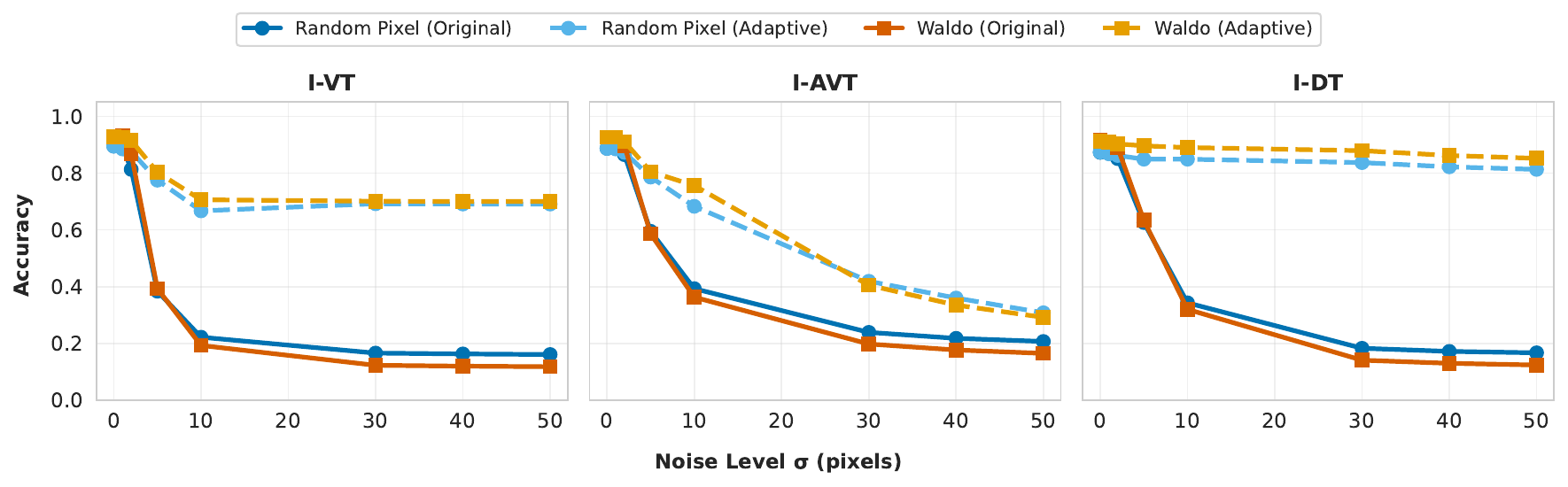}
\hfill
\caption{Robustness analysis of fixation classification algorithms under Gaussian spatial noise ($\sigma = 0$-$50$ pixels). \textbf{Top row:} Evolution of adaptive thresholds for I-VT, I-AVT, and I-DT algorithms across noise levels for Random Pixel and Waldo tasks. \textbf{Bottom row:} Classification accuracy comparing fixed thresholds (optimized at $\sigma = 0$) versus adaptive thresholds (re-optimized at each noise level). Solid lines represent original (fixed) thresholds; dashed lines represent adaptive thresholds. Noise amplitude \(\sigma\) is specified in screen pixels (px), since perturbations are applied directly to the recorded (x,y) pixel coordinates.}
\label{fig:noise_analysis}
\end{figure}

The effect of added Gaussian noise ($\sigma$ = 0-50 pixels) was examined both with the original threshold (rapid degradation) -- cf. Tabs.~\ref{tab:noise_fixed_RP}-\ref{tab:merged_idt} in the supplementary section -- and with thresholds re-optimized at each noise level via the $K$-ratio. See Fig.~\ref{fig:noise_analysis}. Under clean recording conditions, I-VT yields the highest agreement with the reference parser. When spatial noise is present, I-VT and I-AVT showed a progressive drift toward saccade-dominated classifications as noise increased, driven by inflated velocity and angular-velocity estimates. I-DT preserved more stable fixation proportions at intermediate noise levels but also deteriorated at higher amplitudes. To quantify the corresponding change in event structure, we measured fixation-duration distributions under noise for I-VT with fixed thresholds the median fixation duration collapses from 619~ms at $\sigma=0$ to 1~ms for $\sigma\ge5$ (with $\Pr[T_{\mathrm{fix}}\ge200~\mathrm{ms}]=0$ from $\sigma\ge2$), whereas adaptive I-DT retains substantially longer fixations, with median 318~ms at $\sigma=0$ and 431~ms at $\sigma=50$ and $\Pr[T_{\mathrm{fix}}\ge200~\mathrm{ms}]$ increasing from 0.887 ($\sigma=0$) to 0.929 ($\sigma=50$), consistent with fixation stretching via saccade absorption.
With adaptive thresholding, I-DT shows the greatest stability in event proportions, while velocity-based algorithms remain markedly more sensitive to recording precision. However, in all cases, the $K$-ratio-based threshold at least partially mitigated against noise.

\section{Discussion and conclusion}\label{sec12}

This study compared three threshold-based classifiers, I-VT, I-AVT, and I-DT, on free viewing (Random Pixel) and visual search (Waldo) eye-tracking data, using EyeLink labels as a reference. Because the EyeLink parser is itself a threshold-based classifier, our reported accuracies quantify agreement with a widely used commercial event-detection pipeline, not an absolute physiological ground truth. Across both tasks, I-VT generally showed high agreement with the benchmark and the most favorable balance between fixation and saccade classification. 
Task structure influenced the optimal thresholds obtained by $K$-ratio minimization. The Waldo dataset required higher velocity and angular-velocity thresholds than the Random Pixel dataset, which reflects more variable and goal-directed oculomotor dynamics in visual search. At the same time, both datasets were dominated by fixations, with saccades forming only a small fraction of samples. This class imbalance means that accuracy alone is not a sufficient measure of performance. This is evident in noise analysis, where adaptive I-DT maintains high overall accuracy largely by preserving fixation performance, while saccade detection (particularly recall) degrades. The class-wise metrics in the Results show that a classifier can achieve relatively high accuracy while failing to identify many saccades. This supports the use of F1-scores, confusion matrices, and explicit event proportions, in line with earlier discussions of evaluation practice in eye movement classification \cite{zemblys}.

The $K$-ratio criterion provided a simple, task-specific method to tune thresholds without explicit modelling of the underlying velocity distributions. By minimizing the discrepancy between empirical and independent transition probabilities, it favored segmentation in which fixations and saccades formed temporally coherent states. This approach is complementary to more elaborate state-space models, such as hidden Markov models, which require stronger assumptions about observation statistics \cite{10.1145/355017.355028}. Here, the $K$-ratio produced consistent participant-specific thresholds and offered a transparent way to adapt classical I-VT, I-AVT, and I-DT schemes to raw eye-tracking data. The analysis of individual differences supports the adaptive interpretation of the method: the $K$-ratio minimum is stable enough to estimate at the participant level, yet variable enough across participants that individualized threshold selection is justified.
Notably, the repeatability of the minimum $K$-ratio itself was consistently higher than the repeatability of the threshold location. This indicates that the optimization criterion is robust even when the exact threshold achieving the minimum shifts somewhat across trials.

The noise analysis confirmed that all three algorithms are sensitive to measurement noise. Even modest Gaussian perturbations led to a marked reduction in agreement with the EyeLink labels and to substantial shifts in event composition. In all cases, re-optimizing thresholds by $K$-ratio mitigated, but did not entirely prevent, the loss of performance. These patterns are consistent with the intended effect of transition minimization, which stabilizes segmentation under noise by prioritizing temporal coherence, but can reduce sensitivity to short saccades. These findings agree with previous work emphasizing the importance of careful preprocessing and denoising in eye-tracking analysis \cite{niehorster2020noise,niehorster2017denoising,LARSSON2015145}. From an applications perspective, the results suggest that I-VT can serve a reasonable default when high-quality data are available, and the goal is to obtain a reliable first-order segmentation of fixations and saccades. I-AVT, despite its directional component, offered limited benefit in this context because of its strong under-detection of saccades. I-DT may still be useful in settings where long, stable fixations are of primary interest, such as some reading and static-scene paradigms \cite{68d3bfb814cc4a61aae0131dd135c9ec}, but it is less suitable for dynamic visual search. In all cases, practitioners should be aware that fixed thresholds calibrated on one task or recording setup may not transfer to others, and that auxiliary events such as microsaccades, smooth pursuits, and glissades remain largely unmodeled \cite{martinezs}.

There are several natural directions for future work. One is to combine $K$-ratio tuning with more advanced denoising and interpolation schemes, which could extend the useful operating range of classical algorithms in noisy or mobile settings. Another is to embed $K$-ratio or related transition statistics into learning-based classifiers, for example, as regularization terms or diagnostic tools for model calibration \cite{FRANCESCHIELLO2022106929,zemblys}. A third is to pursue more systematic study of inter-individual differences, with thresholds optimized at the individual level, to clarify how much of the observed variability is due to genuine oculomotor differences and how much is an artifact of algorithm choice \cite{HESSELS2019100710}. Because both our methods and the EyeLink parser are rule-based detectors, the reported accuracies quantify agreement with a standard commercial parser, which validates against expert manual coding \cite{Andersson2017OneAlgorithm} or synthetic ground-truth data remains an important direction for future work.

In conclusion, I-VT, I-AVT, and I-DT remain useful tools for eye movement classification, but their performance depends strongly on task characteristics, threshold selection, and data quality. The $K$-ratio provides a simple and general way to adapt these methods to new datasets by minimizing unnecessary state transitions. Researchers should choose algorithms and thresholds in light of specific analytical goals, apply robust preprocessing, and be cautious in interpreting fine-grained differences in eye movement metrics when classification uncertainty is high. The development of standardized analysis protocols \cite{5523936} remains an important goal for the field.

\backmatter
\section*{Competing interests} 
The authors declare no competing interests.
\section*{Acknowledgements}

The authors gratefully acknowledge support from the Research Council of Norway through the ``VirtualEye'' project (Ref. 335940-FORSKER22).

\bibliography{sn-bibliography}

\begin{appendices}

\newpage
\section{Assumption checks for the K-ratio framework}
\begin{figure}[h]
\centering
\includegraphics[width=\textwidth]{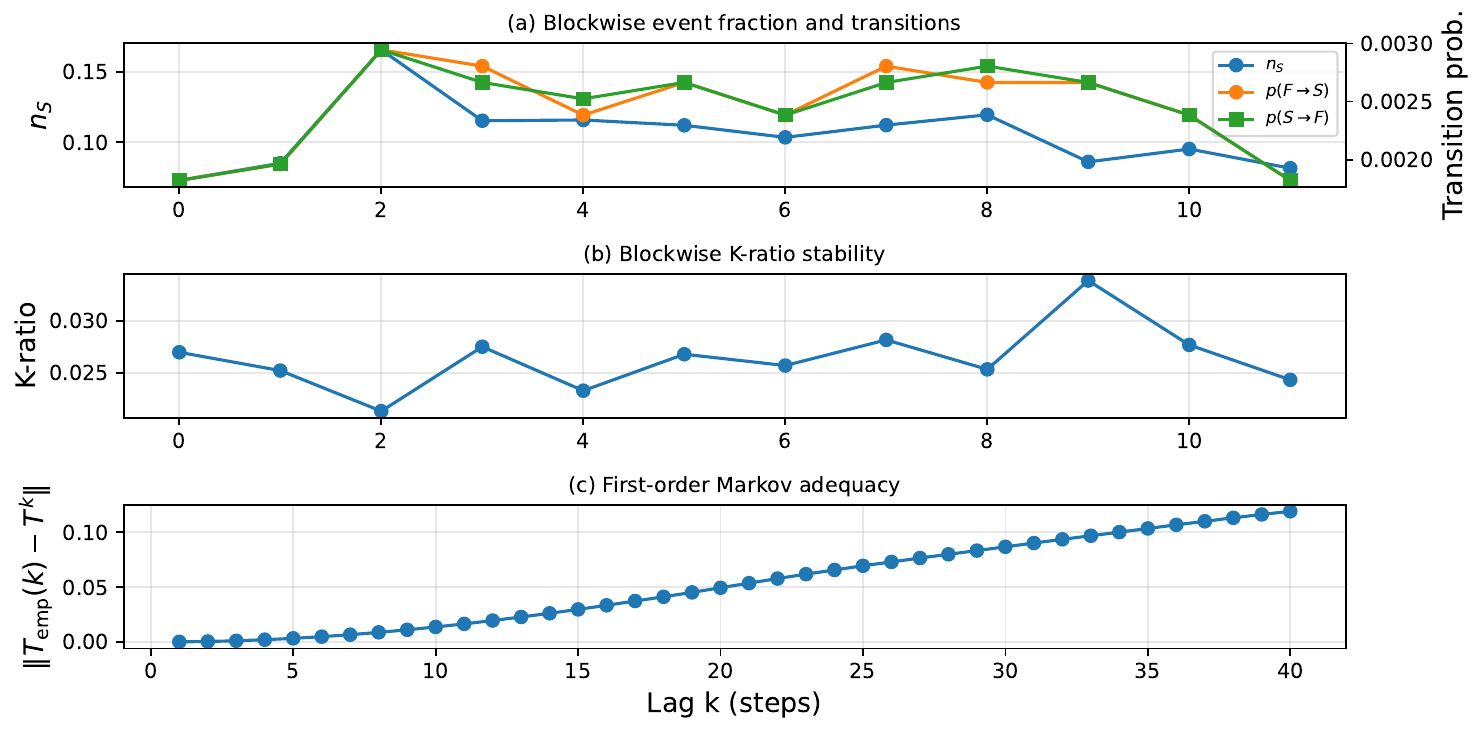}
\caption{Stationarity and Markov adequacy diagnostics. (a) Blockwise \(n_S\), \(p(F\to S)\), \(p(S\to F)\). (b) Blockwise (K)-ratio stability. (c) Deviation \(|T_{\mathrm{emp}}(k)-T^k|\) as a function of lag \(k\), showing very small error at short lags.}
\label{markov}
\end{figure}

To examine whether the stationarity assumption used to express the $K$-ratio in empirical form (Eq.~\eqref{k_ratio}) is reasonable for our tasks, we performed two diagnostics on the reference fixation: saccade label sequence. First, we divided each recording of the Random Pixel task (3 trials, each of 60 seconds) into 12 equal-duration, non-overlapping temporal blocks, and computed within each block the saccade fraction $n_S$, the one-step transition probabilities $p(F\to S)$ and $p(S\to F)$, and the corresponding blockwise $K$-ratio. This provides a direct check for slow within-task drift in event composition and switching. In the Random Pixel example shown in Fig.~\ref{markov}, $n_S$ varies modestly across blocks ($0.073$--$0.166$). At the same time, transition probabilities remain close ($p(F\to S)\in[0.00182,0.00294]$ and $p(S\to F)\in[0.00182,0.00294]$ per step). The $K$-ratio stays in a narrow range ($K\in[0.021,0.034]$, mean $0.026\pm0.003$), indicating that the transition-minimization criterion is stable under mild nonstationarity. Second, we evaluated short-lag Markov adequacy by comparing empirical $k$-step transition matrices to the first-order Markov prediction $T^k$. The mean absolute deviation $\|T_{\mathrm{emp}}(k)-T^k\|$ is near zero at short lags (e.g., $3.35\times10^{-4}$ at $k=2$, $9.98\times10^{-4}$ at $k=3$, and $3.21\times10^{-3}$ at $k=5$), supporting a first-order proxy at the timescale most relevant for suppressing spurious switching and selecting thresholds via $K$-ratio minimization. These checks do not assert that gaze dynamics are globally stationary; rather, they show that a first-order Markov approximation is an adequate local model for the purpose of threshold tuning in our datasets.

In addition to the visual blockwise diagnostics, we performed a regression-based trend check across blocks. For $n_S$, the linear trend was not significant ($p=0.556$), and the quadratic component was marginal ($p=0.060$), while for the $K$-ratio, neither a linear trend ($p=0.278$) nor a quadratic component ($p=0.941$) was significant. This supports the interpretation that any slow drift is modest and does not invalidate a first-order Markov proxy for local threshold tuning.

\section{Stability of participant-specific threshold, and  $K$-ratio across trials}\label{app:icc}

\begin{figure*}[h]
\centering
\includegraphics[width=0.49\textwidth]{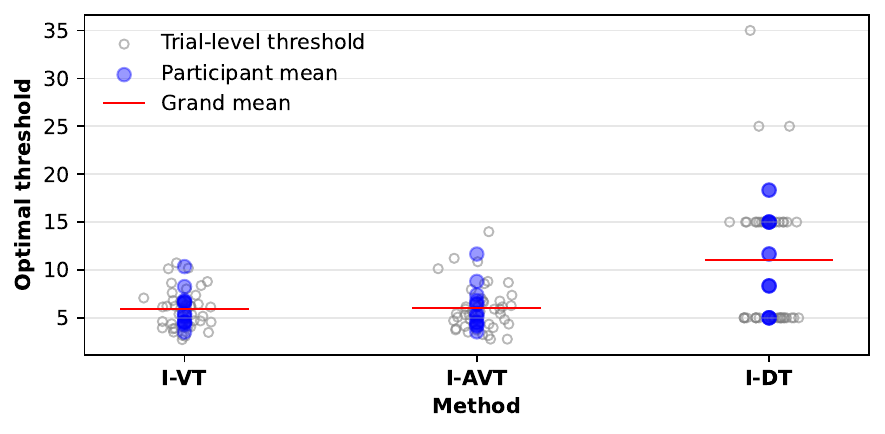}
\includegraphics[width=0.49\textwidth]{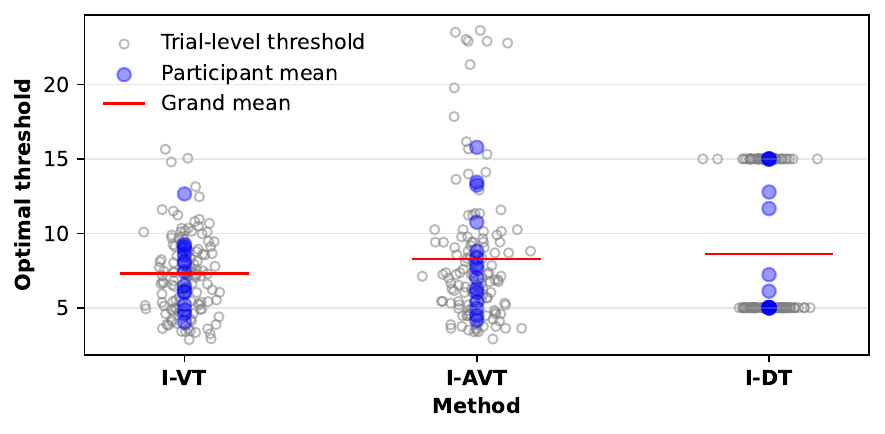}
\caption{Distribution of participant-specific optimal thresholds obtained by minimizing the $K$-ratio, shown separately for I-VT, I-AVT, and I-DT. Gray open circles represent trial-level optimal thresholds, blue filled circles indicate participant-wise mean thresholds across trials, and red horizontal lines mark the grand mean across all observations for each method.}
\label{fig:icc_threshold}
\end{figure*}
To examine whether the $K$-ratio minimum reflects a stable participant-specific property rather than only a pooled task-level optimum, we repeated the threshold optimization separately for each participant and each trial. For the Random Pixel task, optimization was performed for the three 60\,s trials. For the Waldo task, optimization was performed for the nine 45\,s trials. This yielded, for each participant and each trial, an optimal threshold for I-VT, I-AVT, and I-DT together with the corresponding minimum $K$-ratio value.

\begin{figure*}[t]
\centering
\includegraphics[width=\textwidth]{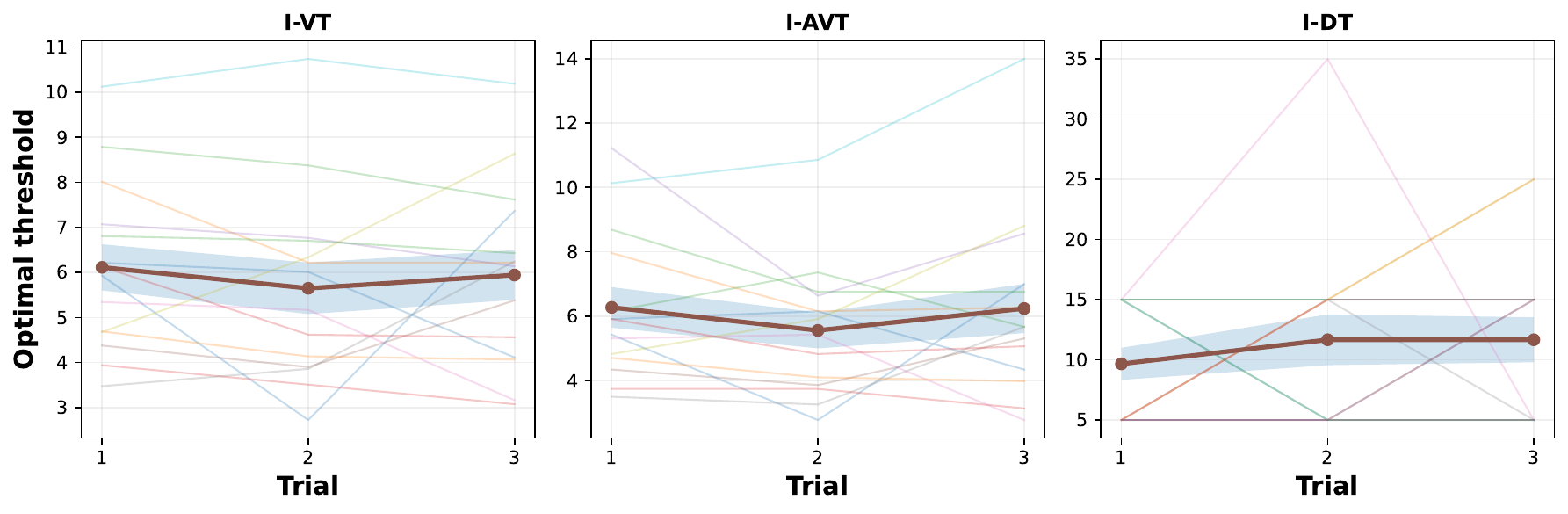}
\includegraphics[width=\textwidth]{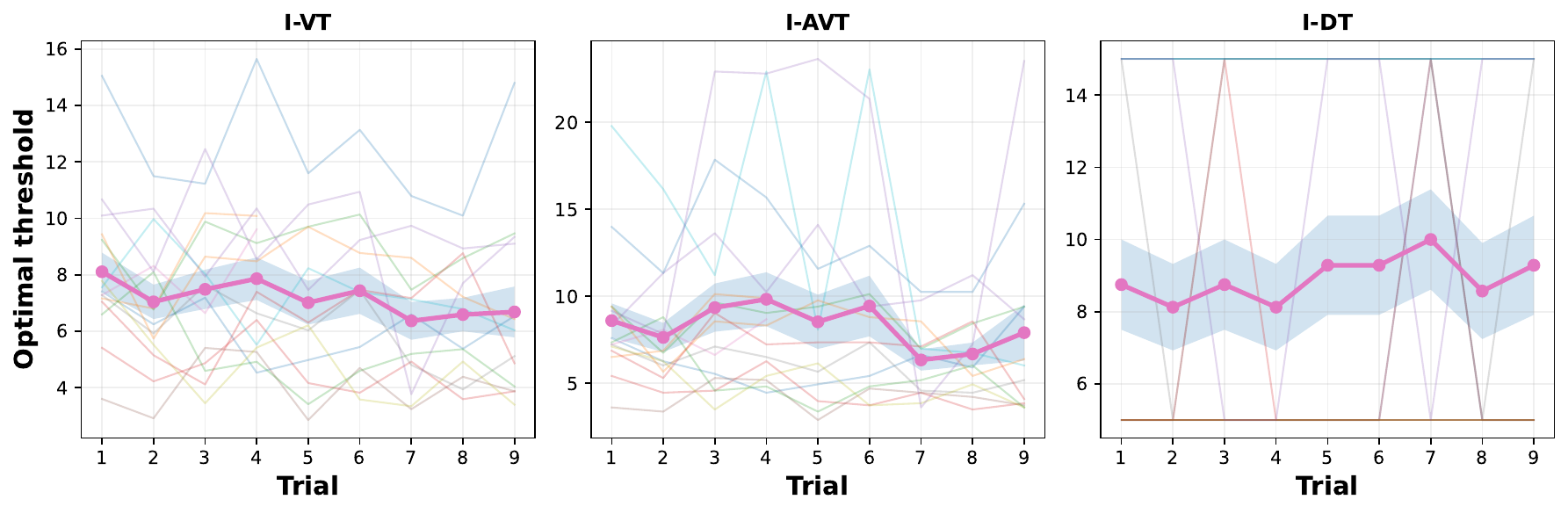}
\caption{Trial-wise participant-specific thresholds obtained by minimizing the $K$-ratio. Top row: Random Pixel. Bottom row: Waldo. Panels show I-VT, I-AVT, and I-DT separately. Thin colored lines correspond to individual participants. The thick dark line shows the mean threshold across participants at each trial, and the shaded region indicates the standard error of the mean.}
\label{fig:icc_thresholds}
\end{figure*}

The resulting trial-wise thresholds are shown in Fig.~\ref{fig:icc_threshold} and Fig.~\ref{fig:icc_thresholds}. For each measure $y_{ij}$, where $i$ indexes participant and $j$ trial, repeatability across trials was quantified using the one-way random-effects intraclass correlation coefficient, ICC$(1,1)$. Under the variance-components model
\begin{equation}
y_{ij}=\mu+a_i+\varepsilon_{ij},
\end{equation}
with between-participant variance $\sigma_p^2$ and within-participant variance $\sigma_t^2$, the ICC is
\begin{equation}
\mathrm{ICC}=\frac{\sigma_p^2}{\sigma_p^2+\sigma_t^2}.
\end{equation}
In practice, ICC$(1,1)$ was estimated from the ANOVA mean squares as
\begin{equation}
\mathrm{ICC}(1,1)=
\frac{MS_{\mathrm{between}}-MS_{\mathrm{within}}}
{MS_{\mathrm{between}}+(k-1)MS_{\mathrm{within}}},
\end{equation}
where $MS_{\mathrm{between}}$ and $MS_{\mathrm{within}}$ are the between- and within-participant mean squares, respectively, and
\begin{equation}
k=\frac{n}{\sum_{i=1}^{n} 1/k_i}
\end{equation}
is the harmonic mean number of valid trials per participant for the unbalanced design. Statistical significance was assessed with the standard ANOVA ratio
\begin{equation}
F=\frac{MS_{\mathrm{between}}}{MS_{\mathrm{within}}},
\end{equation}
with degrees of freedom $df_1=n-1$ and $df_2=N-n$, where $n$ is the number of participants and $N=\sum_i k_i$ the total number of observations.

The ICC results are summarized in Table~\ref{tab:icc_kratio}. In the Random Pixel task, the participant-specific optimal thresholds were moderately repeatable across trials for I-VT and I-AVT, with ICC values of 0.688 and 0.676, respectively. In contrast, the I-DT spatial threshold showed substantially weaker repeatability, with ICC = 0.257. The minimum $K$-ratio values at the participant-specific optima were more repeatable than the thresholds themselves, with ICC values of 0.816 for I-VT, 0.808 for I-AVT, and 0.779 for I-DT.

In the Waldo task, repeatability was generally stronger. The I-VT threshold yielded ICC = 0.728, the I-AVT threshold ICC = 0.544, and the I-DT spatial threshold ICC = 0.808. The minimum $K$-ratio values again showed high repeatability, with ICC values of 0.853 for I-VT, 0.845 for I-AVT, and 0.859 for I-DT. All Waldo ICC estimates were highly significant. In the Random Pixel task, all intraclass correlations were significant except for the case of the I-DT threshold, which was only marginal ($p = 0.050$).
\begin{figure*}[t]
\centering
\includegraphics[width=\textwidth]{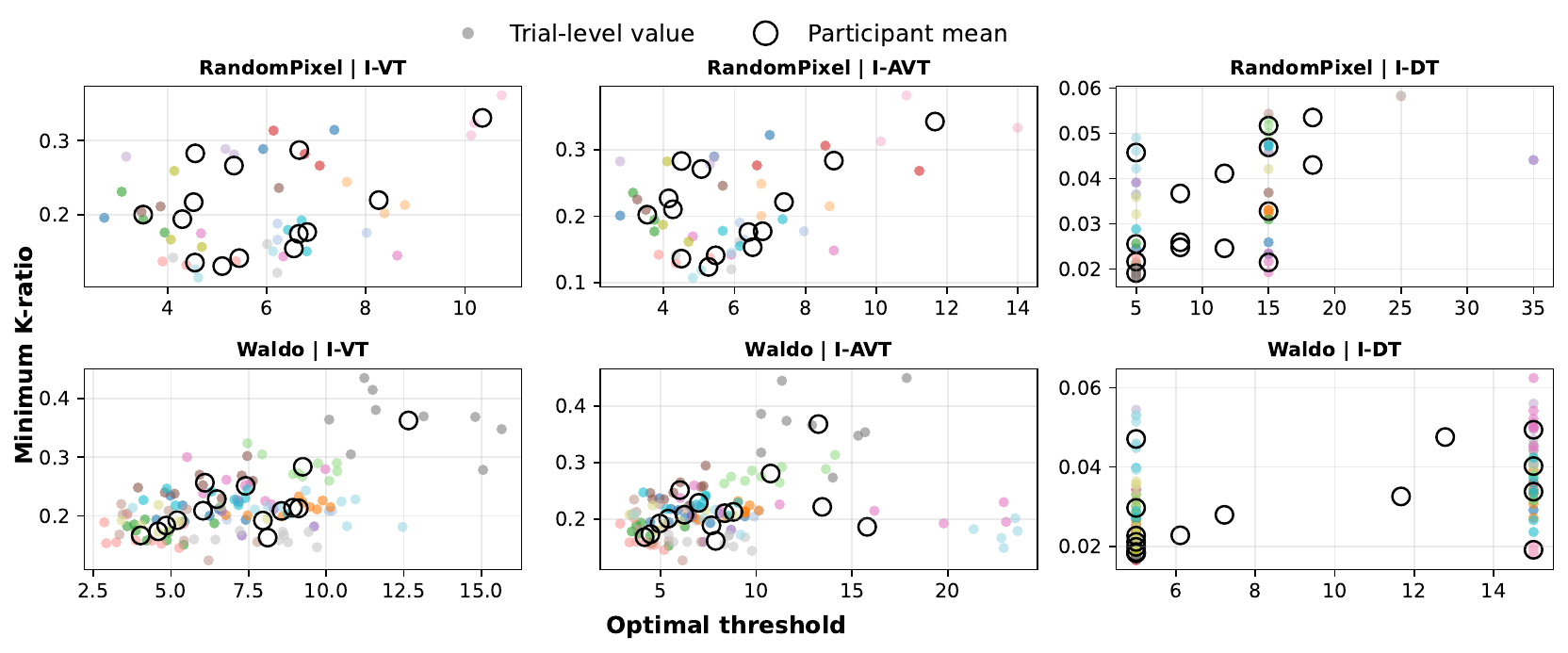}
\caption{Participant-specific optimal thresholds estimated separately for each trial by minimizing the $K$-ratio. Colored dots represent trial-level values and black open circles represent participant means.}
\label{fig:kratio_thresholds}
\end{figure*}

These findings support two related conclusions. First, the threshold that minimizes the $K$-ratio is not merely a pooled artifact. It shows measurable within-participant consistency across trials, indicating that the optimization is capturing stable structure in the eye-movement sequence. Second, the thresholds are not identical across participants. This between-participant variation is visible in Fig.~\ref{fig:icc_thresholds} and is reflected statistically by high ICC values.

\begin{table*}[t]
\centering
\caption{Repeatability across trials of participant-specific optimal thresholds and corresponding minimum $K$-ratio values, quantified using one-way random-effects intraclass correlation, ICC$(1,1)$. A higher ICC indicates greater within-participant consistency across trials.}
\label{tab:icc_kratio}
\begin{tabular}{@{}llccccc@{}}
\toprule
\textbf{Task} & \textbf{Measure} & \textbf{ICC$(1,1)$} & \textbf{$F$}  & \textbf{$p$} \\
\midrule
RPixel & I-VT threshold & 0.688 & 7.613  & $3.92 \times 10^{-6}$ \\
RPixel & I-AVT threshold & 0.676 & 7.257  & $6.24 \times 10^{-6}$ \\
RPixel & I-DT threshold  & 0.257 & 2.036  & 0.0502 \\
RPixel & $K$-ratio at I-VT optimum & 0.816 & 14.296  & $4.58 \times 10^{-9}$ \\
RPixel & $K$-ratio at I-AVT optimum & 0.808 & 13.621 & $7.97 \times 10^{-9}$ \\
RPixel & $K$-ratio at I-DT optimum & 0.779 & 11.603 & $1.64 \times 10^{-8}$ \\
\midrule
Waldo & I-VT threshold & 0.728 & 21.695 &  $< 10^{-15}$ \\
Waldo & I-AVT threshold & 0.544 & 10.216 & $2.50 \times 10^{-14}$ \\
Waldo & I-DT threshold  & 0.808 & 33.854 & $< 10^{-15}$ \\
Waldo & $K$-ratio at I-VT optimum & 0.853 & 45.729 & $< 10^{-15}$ \\
Waldo & $K$-ratio at I-AVT optimum & 0.845 & 43.072 & $< 10^{-15}$ \\
Waldo & $K$-ratio at I-DT optimum & 0.859 & 48.274 & $< 10^{-15}$ \\
\bottomrule
\end{tabular}
\end{table*}

\section{Noise robustness of fixed and adaptive thresholds}
\label{app:noise}

In this supplementary section, we provide detailed results from the noise robustness analysis. Table~\ref{tab:k_ratio_all} reports the task-specific thresholds obtained by $K$-ratio minimization under noise-free conditions. These values served as the fixed (non-adaptive) baseline thresholds in subsequent experiments. 
\begin{table*}[b]
\centering
\caption{$K$-ratio-optimized thresholds for the three algorithms (zero added noise).}
\label{tab:k_ratio_all}
\begin{tabular}{@{}lcccc@{}}
\toprule
\textbf{Algorithm} & \multicolumn{2}{c}{\textbf{Random Pixel}} & \multicolumn{2}{c}{\textbf{Waldo}} \\
\cmidrule(lr){2-3} \cmidrule(lr){4-5}
 & Threshold & $K$-ratio & Threshold & $K$-ratio \\
\midrule
I-VT  & 5.27\,px/s  & 0.189 & 6.57\,px/s  & 0.179 \\
I-AVT & 5.31\,px/s & 0.190 & 6.75\,px/s & 0.181 \\
I-DT  & 10.20\,px   & 0.020 & 4.98\,px    & 0.029 \\
\bottomrule
\end{tabular}
\end{table*}
\begin{table*}[htbp]
\centering
\caption{Classification degradation with fixed thresholds under increasing noise for I-VT algorithm. \textit{Note:} Agr. = agreement with noise level 0 ($\sigma=0$) (\%); Fix/Sac = sample counts; $\Delta$\% = change in fixation proportion from baseline.}
\label{tab:noise_fixed_RP}
\begin{tabular}{@{}lrrrrr|rrrrr@{}}
\toprule
& \multicolumn{5}{c|}{\textbf{Random Pixel}} & \multicolumn{5}{c}{\textbf{Waldo}} \\
\cmidrule(lr){2-6} \cmidrule(l){7-11}
$\sigma$ & Agr. & Fix & Sac & Fix\% & $\Delta$\% & Agr. & Fix & Sac & Fix\% & $\Delta$\% \\
\midrule
0  & 100.0 & 80478 & 5110  & 94.0 & 0.0  & 100.0 & 47895 & 2104  & 95.8 & 0.0 \\
1  & 98.5  & 79814 & 5774  & 93.3 & 0.8  & 99.3  & 47758 & 2241  & 95.5 & 0.3 \\
2  & 80.4  & 64617 & 20971 & 75.5 & 18.5 & 91.4  & 43965 & 6034  & 87.9 & 7.9 \\
5  & 27.9  & 19694 & 65894 & 23.0 & 71.0 & 36.8  & 16809 & 33190 & 33.6 & 62.2 \\
10 & 12.0  & 5605  & 79983 & 6.6  & 87.5 & 13.7  & 5048  & 44951 & 10.1 & 85.7 \\
30 & 6.7   & 708   & 84880 & 0.8  & 93.2 & 5.3   & 596   & 49403 & 1.2  & 94.6 \\
40 & 6.4   & 383   & 85205 & 0.4  & 93.6 & 4.8   & 315   & 49684 & 0.6  & 95.2 \\
50 & 6.2   & 245   & 85343 & 0.3  & 93.7 & 4.6   & 222   & 49777 & 0.4  & 95.4 \\
\bottomrule
\end{tabular}
\end{table*}
\begin{table*}[htbp]
\centering
\caption{I-VT: Original vs.\ Adaptive threshold performance across noise levels. \textit{Note:} Acc = Accuracy; F1\textsubscript{F} = F1-score (Fixation); F1\textsubscript{S} = F1-score (Saccade).}
\label{tab:merged_ivt}
\begin{tabular}{@{}l|ccc|ccc|ccc|ccc@{}}
\toprule
& \multicolumn{6}{c|}{\textbf{Random Pixel}} & \multicolumn{6}{c}{\textbf{Waldo}} \\
\cmidrule(lr){2-7} \cmidrule(l){8-13}
& \multicolumn{3}{c|}{Original} & \multicolumn{3}{c|}{Adaptive} & \multicolumn{3}{c|}{Original} & \multicolumn{3}{c}{Adaptive} \\
\cmidrule(lr){2-4} \cmidrule(lr){5-7} \cmidrule(lr){8-10} \cmidrule(l){11-13}
$\sigma$ & Acc & F1\textsubscript{F} & F1\textsubscript{S} & Acc & F1\textsubscript{F} & F1\textsubscript{S} & Acc & F1\textsubscript{F} & F1\textsubscript{S} & Acc & F1\textsubscript{F} & F1\textsubscript{S} \\
\midrule
0 & 0.90 & 0.94 & 0.51 & 0.90 & 0.94 & 0.51 & 0.93 & 0.96 & 0.54 & 0.93 & 0.96 & 0.54 \\
1 & 0.90 & 0.94 & 0.52 & 0.89 & 0.94 & 0.44 & 0.93 & 0.96 & 0.56 & 0.93 & 0.96 & 0.53 \\
2 & 0.81 & 0.89 & 0.43 & 0.88 & 0.93 & 0.38 & 0.87 & 0.93 & 0.44 & 0.92 & 0.95 & 0.45 \\
5 & 0.38 & 0.46 & 0.29 & 0.78 & 0.87 & 0.28 & 0.39 & 0.50 & 0.22 & 0.80 & 0.89 & 0.25 \\
10 & 0.22 & 0.16 & 0.27 & 0.67 & 0.79 & 0.24 & 0.19 & 0.18 & 0.21 & 0.71 & 0.82 & 0.19 \\
30 & 0.17 & 0.02 & 0.27 & 0.69 & 0.81 & 0.19 & 0.12 & 0.02 & 0.21 & 0.70 & 0.82 & 0.16 \\
40 & 0.16 & 0.01 & 0.27 & 0.69 & 0.81 & 0.19 & 0.12 & 0.01 & 0.21 & 0.70 & 0.82 & 0.16 \\
50 & 0.16 & 0.01 & 0.27 & 0.69 & 0.81 & 0.19 & 0.12 & 0.01 & 0.21 & 0.70 & 0.82 & 0.16 \\
\bottomrule
\end{tabular}
\end{table*}
\begin{table*}[htbp]
\centering
\caption{I-AVT: Original vs.\ Adaptive threshold performance across noise levels. See Table~\ref{tab:merged_ivt}.}
\label{tab:merged_iavt}
\begin{tabular}{@{}l|ccc|ccc|ccc|ccc@{}}
\toprule
& \multicolumn{6}{c|}{\textbf{Random Pixel}} & \multicolumn{6}{c}{\textbf{Waldo}} \\
\cmidrule(lr){2-7} \cmidrule(l){8-13}
& \multicolumn{3}{c|}{Original} & \multicolumn{3}{c|}{Adaptive} & \multicolumn{3}{c|}{Original} & \multicolumn{3}{c}{Adaptive} \\
\cmidrule(lr){2-4} \cmidrule(lr){5-7} \cmidrule(lr){8-10} \cmidrule(l){11-13}
$\sigma$ & Acc & F1\textsubscript{F} & F1\textsubscript{S} & Acc & F1\textsubscript{F} & F1\textsubscript{S} & Acc & F1\textsubscript{F} & F1\textsubscript{S} & Acc & F1\textsubscript{F} & F1\textsubscript{S} \\
\midrule
0 & 0.84 & 0.94 & 0.46 & 0.89 & 0.94 & 0.46 & 0.89 & 0.96 & 0.53 & 0.90 & 0.96 & 0.53 \\
1 & 0.89 & 0.94 & 0.45 & 0.89 & 0.94 & 0.43 & 0.93 & 0.96 & 0.52 & 0.93 & 0.96 & 0.53 \\
2 & 0.87 & 0.92 & 0.40 & 0.87 & 0.93 & 0.36 & 0.90 & 0.94 & 0.43 & 0.91 & 0.95 & 0.42 \\
5 & 0.60 & 0.72 & 0.27 & 0.79 & 0.88 & 0.21 & 0.59 & 0.72 & 0.21 & 0.81 & 0.89 & 0.20 \\
10 & 0.39 & 0.49 & 0.26 & 0.68 & 0.80 & 0.20 & 0.36 & 0.47 & 0.20 & 0.76 & 0.86 & 0.14 \\
30 & 0.24 & 0.21 & 0.27 & 0.42 & 0.52 & 0.26 & 0.20 & 0.20 & 0.20 & 0.41 & 0.53 & 0.19 \\
40 & 0.22 & 0.16 & 0.27 & 0.36 & 0.44 & 0.26 & 0.18 & 0.15 & 0.20 & 0.34 & 0.43 & 0.20 \\
50 & 0.21 & 0.13 & 0.27 & 0.31 & 0.35 & 0.27 & 0.17 & 0.12 & 0.20 & 0.29 & 0.37 & 0.20 \\
\bottomrule
\end{tabular}
\end{table*}
\begin{table*}[htbp]
\centering
\caption{I-DT: Original vs.\ Adaptive threshold performance across noise levels. See Table~\ref{tab:merged_ivt}.}
\label{tab:merged_idt}
\begin{tabular}{@{}l|ccc|ccc|ccc|ccc@{}}
\toprule
& \multicolumn{6}{c|}{\textbf{Random Pixel}} & \multicolumn{6}{c}{\textbf{Waldo}} \\
\cmidrule(lr){2-7} \cmidrule(l){8-13}
& \multicolumn{3}{c|}{Original} & \multicolumn{3}{c|}{Adaptive} & \multicolumn{3}{c|}{Original} & \multicolumn{3}{c}{Adaptive} \\
\cmidrule(lr){2-4} \cmidrule(lr){5-7} \cmidrule(lr){8-10} \cmidrule(l){11-13}
$\sigma$ & Acc & F1\textsubscript{F} & F1\textsubscript{S} & Acc & F1\textsubscript{F} & F1\textsubscript{S} & Acc & F1\textsubscript{F} & F1\textsubscript{S} & Acc & F1\textsubscript{F} & F1\textsubscript{S} \\
\midrule
0 & 0.90 & 0.93 & 0.42 & 0.87 & 0.93 & 0.44 & 0.94 & 0.95 & 0.48 & 0.91 & 0.95 & 0.54 \\
1 & 0.87 & 0.93 & 0.42 & 0.87 & 0.93 & 0.37 & 0.91 & 0.95 & 0.47 & 0.91 & 0.95 & 0.49 \\
2 & 0.85 & 0.92 & 0.40 & 0.86 & 0.92 & 0.31 & 0.89 & 0.94 & 0.43 & 0.90 & 0.95 & 0.36 \\
5 & 0.63 & 0.75 & 0.30 & 0.85 & 0.92 & 0.11 & 0.63 & 0.76 & 0.24 & 0.90 & 0.94 & 0.20 \\
10 & 0.34 & 0.40 & 0.27 & 0.85 & 0.92 & 0.11 & 0.32 & 0.41 & 0.21 & 0.89 & 0.94 & 0.14 \\
30 & 0.18 & 0.07 & 0.27 & 0.84 & 0.91 & 0.08 & 0.14 & 0.07 & 0.21 & 0.88 & 0.94 & 0.09 \\
40 & 0.17 & 0.04 & 0.27 & 0.82 & 0.90 & 0.10 & 0.13 & 0.04 & 0.21 & 0.86 & 0.93 & 0.10 \\
50 & 0.17 & 0.03 & 0.27 & 0.81 & 0.90 & 0.10 & 0.12 & 0.02 & 0.21 & 0.85 & 0.92 & 0.10 \\
\bottomrule
\end{tabular}
\end{table*}
\begin{figure}[htbp]
\centering
\includegraphics[width=\textwidth]{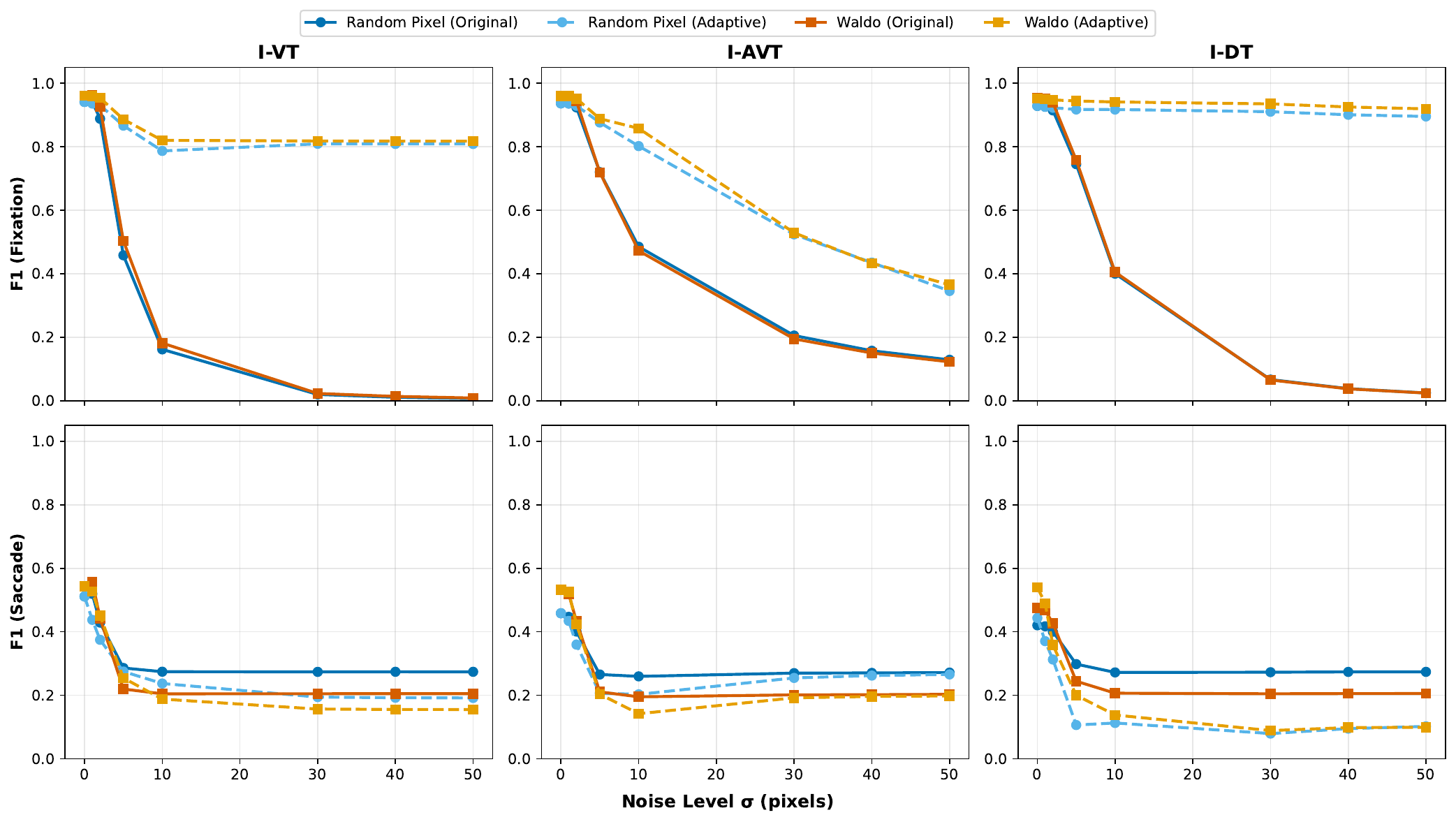}
\caption{F1 scores for fixation and saccade classification across noise levels. Top row: F1-score for fixation detection. Bottom row: F1-score for saccade detection. Solid lines indicate original thresholds; dashed lines indicate adaptive thresholds.}
\label{fig:f1_comparison}
\end{figure}
\begin{figure}[t]
\centering
\includegraphics[width=\textwidth]{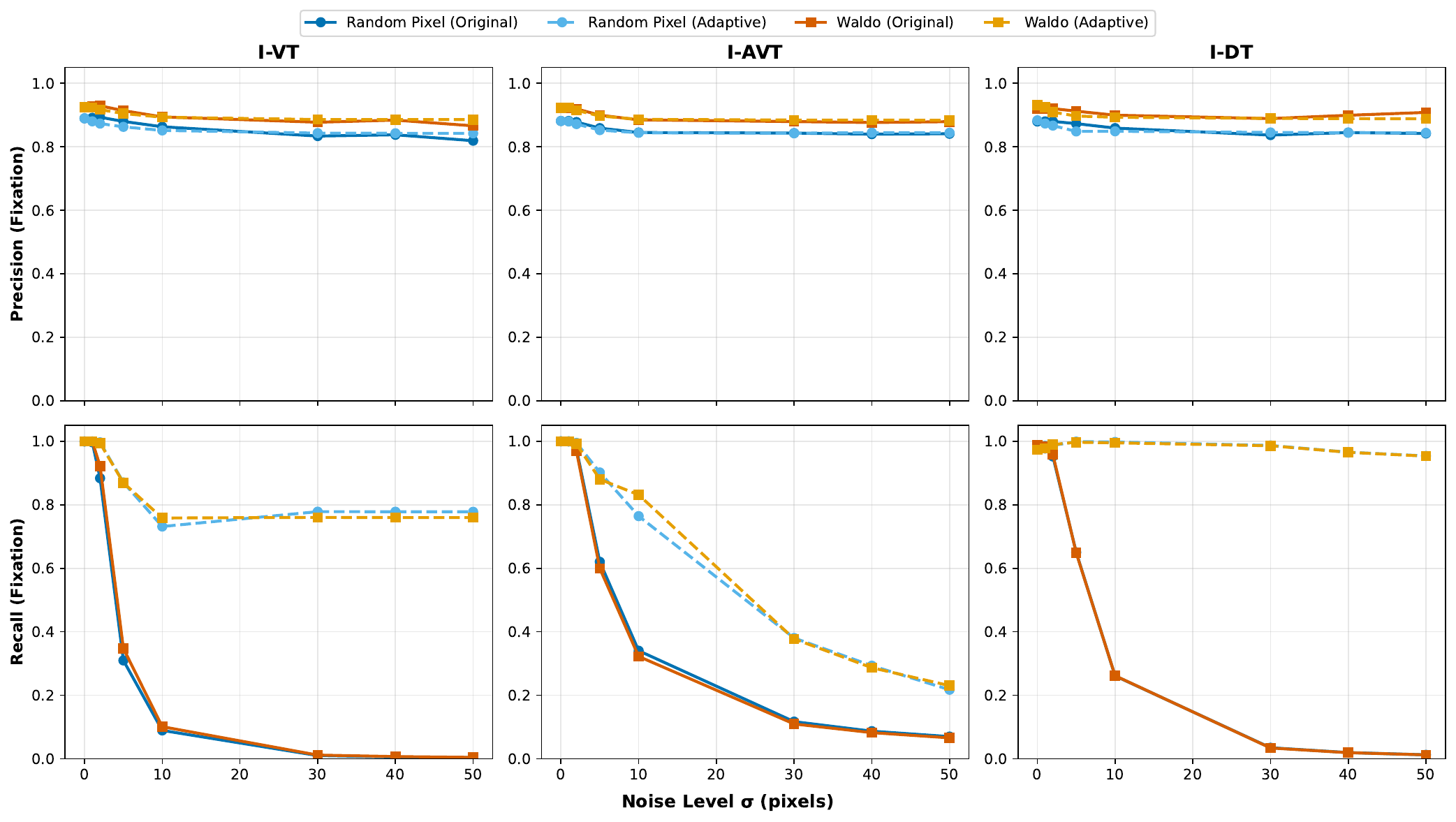}
\caption{Precision and recall for fixation detection across noise levels. Top row: Precision. Bottom row: Recall. Adaptive thresholds maintain higher precision at the cost of reduced recall at extreme noise levels.}
\label{fig:precision_recall}
\end{figure}

Table~\ref{tab:noise_fixed_RP} shows how classification performance degrades 
when optimal noise-free thresholds are held fixed under increasing Gaussian spatial noise ($\sigma = 0$--$50$\,px) for the I-VT algorithm. Agreement with the noise-free classification drops below 90\% at $\sigma = 2$\,px for Random Pixel, while Waldo maintains agreement above 90\% until $\sigma = 5$\,px. At high noise levels ($\sigma \geq 30$\,px), both tasks approach chance-level agreement 
($\sim$5--7\%), with nearly all samples misclassified as saccades.

Tables~\ref{tab:merged_ivt}-\ref{tab:merged_idt} compare accuracy and F1 
scores between fixed thresholds (optimized at $\sigma=0$) and adaptive 
thresholds (re-optimized at each noise level) for all three algorithms 
across both tasks. I-DT with adaptive thresholds demonstrates the greatest robustness, maintaining accuracy above 0.81 even at $\sigma = 50$\,px, whereas I-AVT shows the steepest degradation, falling to 0.31 accuracy at extreme noise levels.

Figure~\ref{fig:f1_comparison} visualizes F1 scores for fixation and saccade detection across noise levels. Adaptive thresholds (dashed lines) preserve fixation F1 scores substantially better than fixed thresholds (solid lines), with I-DT maintaining fixation F1 $>$ 0.90 across all noise levels. However, saccade F1 scores decline for all algorithms under both threshold conditions, with adaptive I-DT showing particularly low saccade F1 ($<$ 0.11) at high noise. Figure~\ref{fig:precision_recall} reveals the precision-recall trade-off underlying adaptive threshold optimization. Precision for fixation detection remains high across noise levels under adaptive thresholding, but recall decreases substantially at $\sigma > 10$\,px. This pattern indicates that adaptive thresholds achieve overall accuracy by conservatively classifying uncertain samples as fixations.

The results demonstrate several key patterns regarding noise robustness. 
Algorithm-specific differences in noise resilience are evident: I-DT with 
adaptive thresholds maintain the highest accuracy even at extreme noise 
conditions, while I-AVT shows the steepest degradation, suggesting that 
angular velocity features are particularly susceptible to noise-induced 
perturbations. A clear precision-recall trade-off emerges with adaptive 
thresholds, while adaptive optimization preserves fixation F1 scores, 
This comes at the cost of saccade detection performance. This strategy 
may be acceptable in applications prioritizing fixation identification, 
but problematic when accurate saccade detection is required.

\end{appendices}

\end{document}